\documentclass{article}

\usepackage[english]{babel}
\usepackage[letterpaper,top=2cm,bottom=2cm,left=3cm,right=3cm,marginparwidth=1.75cm]{geometry}
\usepackage{amsmath}
\usepackage{graphicx}
\usepackage[colorlinks=true, allcolors=blue]{hyperref}
\usepackage{natbib}
\usepackage{xcolor}
\usepackage[utf8]{inputenc}
\usepackage{textgreek}
\usepackage{booktabs}
\usepackage{threeparttable}
\usepackage{multirow}
\usepackage{float}
\usepackage{subcaption}
\usepackage{geometry}
\usepackage{authblk} 
\usepackage{tabularx} 

\title{Bounded Normative Equivalence in Human–AI Cooperation: Group Behaviour, Not Partner Labels, Predicts Cooperation under Anonymous Aggregate Feedback}

\author[1]{Nico Mutzner}
\author[2]{Taha Yasseri}
\author[1]{Heiko Rauhut}

\affil[1]{Department of Sociology, University of Zurich, Zurich, Switzerland}
\affil[2]{Centre for Sociology of Humans and Machines (SOHAM), Trinity College Dublin \& Technological University Dublin, Dublin, Ireland}

\begin{document}
\maketitle

\begin{abstract}
The introduction of artificial intelligence (AI) agents into human group settings raises essential questions about how these novel participants influence cooperative social norms. While prior work has examined human–AI and human–robot teaming in small groups, less is known about whether an AI label alters cooperation and norm-related outcomes in repeated group interactions. This study addresses this gap through an online experiment using a repeated four-player Public Goods Game (PGG). Each group consisted of three human participants and one bot, which was framed either as human or AI and followed one of three predefined decision strategies: unconditional cooperation, conditional cooperation, or free-riding. In our sample of 236 participants, cooperation was primarily associated with how much the group had contributed in the previous round and how much participants themselves had contributed previously. These patterns were similar across human- and AI-labelled conditions, and cooperation levels did not differ significantly by agent label; a formal equivalence test (TOST) indicated that any label effect on contributions was smaller than ±5 tokens (5\% of the endowment). Furthermore, we found no evidence of label-based differences in norm persistence in a follow-up Prisoner’s Dilemma or in participants’ normative perceptions. Observed cooperation thus followed the same behavioural regularities across conditions, consistent with participants responding to group behaviour rather than to the partner’s label. We describe this pattern as bounded normative equivalence: under anonymous aggregate group feedback, an AI label produced no detectable differences in observed cooperation or norm-related outcomes. We argue that this equivalence is bounded by the informational structure of the setting: aggregate feedback makes individual actions difficult to attribute, diluting the identity cues that might otherwise trigger differentiation. These findings suggest that, in collective settings where individual contributions are not identifiable, cooperative norms can extend to groups that include artificial agents.

\end{abstract}

\section{Introduction}
% Context & Motivation
Finding solutions to climate change, geopolitical instability, and widening resource inequality represents a fundamental challenge: addressing social problems involving public goods often requires cooperation among actors with heterogeneous incentives and information. At the same time, rapid advances in artificial intelligence are reshaping how people solve a wide
range of problems by inserting autonomous agents into domains once reserved for humans. 9 out of 10 organisations already report the regular use of AI in their operations \citep{mckinseyreportStateAI20252025} and two-thirds of people worldwide believe that AI products and services will significantly impact daily life within the next three to five years \citep{StanfordAIIndex2025}. In this study, the term artificial intelligence (AI) is used broadly to refer to autonomous or semi-autonomous decision-making agents that perform tasks or make strategic choices that typically require human judgment. This conceptualization encompasses AI systems that participate in shared decision processes, such as software agents allocating energy across smart grids, algorithms coordinating traffic or resource flows, or automated trading systems making cooperative or competitive moves in markets. As these agents are increasingly embedded in joint decision-making situations, understanding how humans cooperate with and respond to AI partners has become an essential social question \citep{tsvetkovaNewSociologyHumans2024}.

% human–AI Cooperation
Research shows that humans often treat AI as social actors, and that willingness to cooperate with AI depends on framing, perceived warmth, competence, and trust \citep{nassComputersAreSocial1994, demeloImpactEmotionDisplays2011, mckeeWarmthCompetenceHumanagent2024,chongHumanConfidenceArtificial2022, zhangTrustAIHuman2023}. This tendency aligns with the Social Heuristics Hypothesis \citep{randSocialHeuristicsShape2014}, which posits that cooperation is often an intuitive, automated response generalized from daily social interactions. If humans default to these internalized cooperative scripts, they may extend them to artificial partners unless motivated to deliberate. At the same time, studies consistently find that cooperation with AI partners tends to be lower than with human counterparts, a phenomenon recently formalised as the 'machine penalty' \citep{makoviRewardsPunishmentsHelp2025}. This pattern is often linked to algorithm aversion or the strategic exploitation of AI agents perceived as less sensitive to social cues and sanctions than humans \citep{dietvorstAlgorithmAversionPeople2015, karpusAlgorithmExploitationHumans2021, bazaziAIsAssignedGender2025}. Much of the behavioural economics literature on human–AI cooperation remains dyadic, emphasising partner-to-partner reciprocity, whereas adjacent work on human–AI and human–robot teams has examined team composition, trust propagation, and collective intelligence in richer collaborative tasks \citep{mcneeseWhoWhatMy2021, guoTIPTrustInference2024, westbyCollectiveIntelligenceHumanAI2023}. Group structures can diffuse attention and accountability, highlight social influence, and create shared expectations that shape behaviour. Accordingly, the open question is not simply whether people cooperate with AI, but whether mechanisms that sustain group cooperation change when one member is perceived as an AI \citep{reineckeNeedEmpiricalResearch2025, engPeopleTreatSocial2023}.

% Social Norms
To address this question, we turn to social norms: the shared expectations that guide behaviour and sustain cooperation in groups. In repeated group interaction, individuals adapt conditionally to others, aligning choices with empirical expectations (what others do) and injunctive expectations (what others think one ought to do), two expectation types that jointly define a social norm \citep{bicchieriSocialNorms2018,youngEvolutionSocialNorms2015,kolleInfluenceEmpiricalNormative2021}. When these coincide, cooperation is sustained; when they diverge, cooperation erodes \citep{fehrSocialNormsHuman2004, baronchelliShapingNewNorms2024}. Classic experiments in public-goods settings show that individuals tend to condition their contributions on others' contributions, increasing cooperation when others contribute and withdrawing when they do not, an effect known as conditional cooperation \citep{fischbacherArePeopleConditionally2001, keserConditionalCooperationVoluntary2000, thoniConditionalCooperationReview2018}. These findings highlight that cooperation in groups depends on shared expectations and social influence, mechanisms that may also extend to mixed human–AI groups. Despite this growing small-group literature, there is still limited behavioural evidence on human–AI cooperation through the lens of social norms, especially in repeated public-goods settings where participants respond to aggregate group-level feedback rather than identifiable individual actions \citep{reineckeNeedEmpiricalResearch2025}.

% AI Norm Influence
Building on this human evidence, recent research has begun to examine how artificial agents themselves participate in or influence these normative dynamics. AI can shape norms positively by reinforcing fairness, reciprocity, and trust \citep{taddeoHowAICan2018, mccannonArtificialIntelligenceProsocial2024} or negatively by normalizing free-riding and moral disengagement \citep{kobisBadMachinesCorrupt2021, engPeopleTreatSocial2023}. Computational and multi-agent studies likewise show how artificial agents can seed, amplify, or stabilise cooperative norms \citep{shiEnhancingSocialCohesion2024, kulkarniLearningNudgesConditional2024, renEmergenceSocialNorms2024, hintzePromotingCooperationPublic2024}. Together, these strands suggest that AI can mediate, amplify, or dampen the normative forces that sustain cooperation, while leaving open the question of whether norm-guided cooperation changes when one group member is perceived as AI. Yet this work leaves open a more specific question: whether the perceived identity of one group member changes norm-guided cooperation when the interaction structure, incentives, and feedback are held constant. Specifically, it remains unclear whether an AI label alters observable cooperation patterns and norm-related outcomes when the agent is embedded in a small group where feedback is provided at the aggregate level. This creates a theoretical tension: Does an agent's specific identity disrupt social cohesion (differentiation), or are norm-driven contribution patterns robust to labels when participants respond to aggregate group-level feedback?

% Experimental Design
Our central research question is whether the perceived identity of one group member changes cooperation and norm-related cooperation outcomes. To test this, we conducted a between-subjects 2×3 group experiment based on established behavioural-economic games that capture cooperation and norm formation in groups. Participants (three humans and one programmed agent) played ten rounds of a linear Public Goods Game (PGG). The PGG captures how individual incentives conflict with collective welfare, requiring participants to decide how much of their endowment to allocate to a shared group account \citep{chaudhuriSustainingCooperationLaboratory2011,fehrCooperationPunishmentPublic2000}. The fourth player was framed as either human or AI and followed one of three predefined strategies: unconditional cooperator (always contributes the full endowment), conditional cooperator (matches the group’s average contribution from the previous round), or free-rider (always contributes zero). To assess norm persistence beyond the group context, participants then completed a one-shot Prisoner’s Dilemma (PD) with a group partner \citep{nemethCriticalAnalysisResearch1972, doebeliModelsCooperationBased2005, peysakhovichHabitsVirtueCreating2016, stagnaroGoodInstitutionsGenerous2017, arecharExaminingSpilloversLong2018}. Finally, we elicited participants’ perceptions of norms through a coordination measure of social appropriateness for contribution levels \citep{krupkaIdentifyingSocialNorms2013}, along with empirical and injunctive norm expectations. This combined design allows us to test whether the label assigned to one group member changes contribution behaviour, norm perceptions, and subsequent cooperative choices beyond the immediate group context.

% Hypotheses
Based on the theoretical framework outlined above, specific hypotheses were preregistered before data collection (AsPredicted \#234846). Theoretically, the introduction of AI agents creates a tension between two outcomes. On the one hand, a differentiation effect suggests that an agent's artificiality will dampen reciprocity, consistent with algorithm aversion. On the other hand, bounded normative equivalence implies that the aggregate group's behavioural signals may outweigh the agent's perceived identity, such that the label makes little detectable difference to behaviour. Despite this competing possibility, we prioritized the differentiation perspective given the robust evidence of algorithm aversion in dyadic settings. Consequently, we hypothesised lower overall cooperation and weaker normative influence when the fourth member was AI-labelled rather than human-labelled (H1). Beyond this label effect and the groups' collective contribution, we anticipated that the bots' individual behavioural strategy itself would shape cooperation, such that an unconditional cooperator would reinforce cooperative norms (H1a), a conditional cooperator would sustain intermediate levels of cooperation (H1b), and a free-rider would erode them (H1c). Finally, we examined whether cooperative norms established in the group context would persist when participants made subsequent one-on-one decisions. If AI-labelled members evoke weaker social identification and diminished norm pressure, these dynamics should also reduce the internalization and carryover of cooperative norms beyond the group interaction. Accordingly, we predicted that norm persistence would be stronger in human groups compared to human-AI hybrid groups (H2).

% Results & Implications
Results show that group behaviour is the strongest predictor of individual cooperation, regardless of whether the bot is labelled as human or AI, or of its strategy. Differences among the bots' cooperation strategies were small to negligible, and the one-shot Prisoner’s Dilemma revealed no systematic differences by label or strategy concerning norm persistence. Across all conditions, post-task norm perceptions and expectations were closely aligned, and contributions followed the same behavioural regularities: responsiveness to others’ contributions, inertia in individual behaviour, and a gradual decline over time. A formal equivalence test further indicated that any effect of the agent label on contributions was smaller than ±5 tokens (5\% of the endowment). Together, these results point to a pattern of bounded normative equivalence: under anonymous aggregate group feedback, labelling one group member as AI did not detectably alter cooperation or norm-related outcomes. This study advances research on human–AI interaction in two ways. First, it provides group-level evidence for this equivalence in repeated interaction, complementing dyadic work on algorithm aversion. Second, it identifies the informational structure of the setting as a plausible boundary condition: when feedback is aggregated, individual actions are difficult to attribute, diluting the identity cues that might otherwise trigger differentiation. Because many real collective settings share this opacity, this boundary condition is itself informative. Future work should test whether label effects re-emerge when the AI's behaviour is individually identifiable, or when agents are more adaptive, communicative, or socially present. Practical implications for the design of AI systems in teams and institutions suggest that transparent, norm-consistent behaviour may foster social integration more effectively than anthropomorphic design or human-like labelling.
\bigskip

\section{Methods}
\subsection{Experimental Design}
\begin{figure}[H]
  \centering
  \includegraphics[width=1\textwidth]{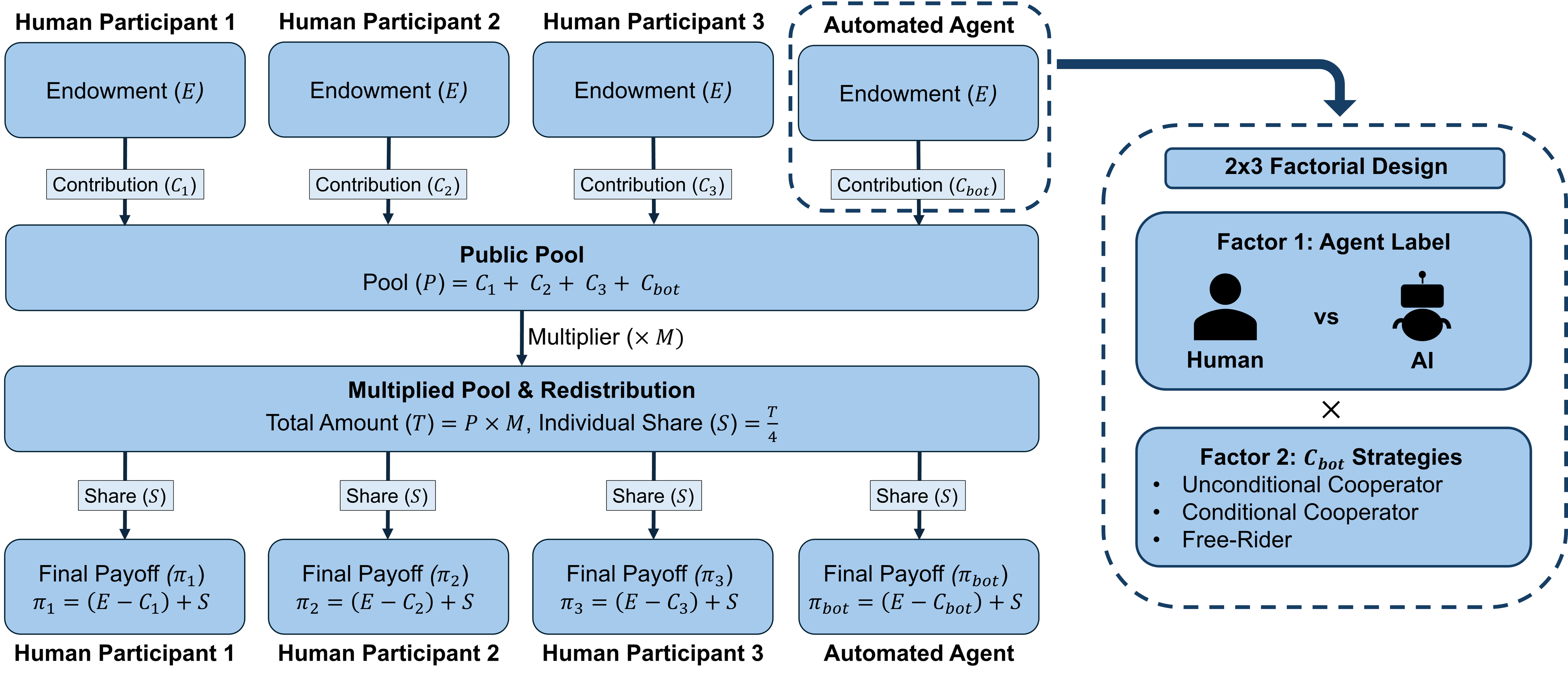}
  \caption{Schematic representation of the experimental design and PGG mechanics}
  \label{fig:Design}
\end{figure}

% Design & Treatment
The experiment employed a between-subjects design, in which each participant played ten rounds of a Public Goods Game (PGG) followed by a single Prisoner’s Dilemma (PD) decision. In the PGG, participants were randomly assigned to one of six experimental conditions in a 2 × 3 factorial design, crossing agent label (human vs AI) with bot strategy (Unconditional Cooperator, Conditional Cooperator, or Free-Rider), see Figure \ref{fig:Design}. In all conditions, groups consisted of three human participants and one computer-controlled player (bot). The interaction structure was identical across treatments; the only differences were (a) the label displayed to participants for the fourth player (human or AI), and (b) the bot’s programmed cooperation strategy. The bot strategies consisted of: 1) Unconditional Cooperator: Always contributes their full endowment in every round, 2) Conditional Cooperator: Matches the average contribution from the previous round, and 3) Free-Rider: Always contributes nothing. The bots' cooperation strategies were not disclosed to the participants. By holding interaction structure constant and varying only the label and cooperation strategy of one agent, we can test whether the normative dynamics of cooperation, namely reciprocity, conformity to group contributions, and norm alignment, remain stable across human and AI conditions. The experiment was implemented in oTree \citep{chenOTreeOpensourcePlatform2016}, an open-source platform for online interactive economic games.

\subsection{Procedure}
Participants first viewed an information page outlining study participation and data protection, followed by an informed consent form. After providing consent, participants received detailed instructions and completed a comprehension test which asked participants to calculate their income for a typical round they would encounter in the game. Participants who failed this test on three consecutive attempts were excluded before the group assignment.
After passing the comprehension check, participants were randomly assigned to four-player groups consisting of three human participants and one computer-controlled player, which was labelled either as human or as AI. Thus, the interaction structure was identical across label conditions; only the displayed identity of the computer-controlled fourth player varied.
Groups then played ten rounds of a linear PGG. In each round, participants received an endowment of 100 tokens and decided how much to contribute to a shared group account. After each round, participants viewed a results page displaying the aggregate group contribution and their individual payoffs. Crucially, to mimic the opacity of large-scale collective action, participants did not see the individual contributions of specific group members. Total contributions were multiplied by 1.5 and evenly redistributed among all four players. Group composition remained constant across rounds. To maintain consistency in group interaction, the session was terminated for all members if any participant dropped out; in such cases, participants were informed that the game had ended and that they would receive only the fixed participation fee.
Following the PGG, participants completed a one-shot Prisoner’s Dilemma (PD) framed as a new interaction with another member of their previous group. In reality, partner responses were always simulated to reflect cooperation, providing an unobtrusive measure of norm persistence after the group interaction.
Finally, participants completed a norm elicitation task, rating the social acceptability of various PGG contribution levels. This was followed by a post-experiment survey measuring perceived trust, fairness, group cohesion, normative pressure, and, for participants in the AI condition, perceptions of the AI player’s role, accountability, fairness, and team membership. The session concluded with a debrief, during which participants reported their perceived group composition (human vs AI) and any suspicions regarding the identity of other group members.

\subsection{Participants}
% Recruitment & Payment
Participants were recruited via Prolific, an established online platform for behavioural research that offers greater participant diversity, more precise screening, and higher data quality than alternative crowdsourcing platforms \citep{peerTurkAlternativePlatforms2017, palanProlificacASubjectPool2018, germineWebGoodLab2012, paolacciTurkUnderstandingMechanical2014}. Five experimental sessions were conducted across three days at varying times to account for potential time-of-day effects. Participants received a fixed payment of £6 per hour plus a performance-based bonus at a rate of 1,500 tokens = £1, resulting in an average effective hourly payment of £8.95, based on the average 17 minutes it took to complete the study. All participants provided informed consent, and the study protocol was approved by the Ethics Committee of University College Dublin (Approval No. 128-LS-C-25-Yasseri).

% Sample Description
A total of 366 individuals entered the study. 37 did not consent, 5 left during the instructions, and 14 failed the comprehension check, leaving 310 who began the experiment. Because the experiment required synchronous four-player groups, attrition by one participant required terminating the session for the remaining group members as well. This design feature amplified the effect of individual dropouts, yielding 246 participants who completed the PGG and 240 who finished the whole experiment. Four additional participants failed the manipulation check, resulting in a final sample of 236. To validate the effectiveness of the group composition manipulation, we asked participants in the post-experiment debrief: "At any point, did you doubt whether the group composition was exactly as described?" Forty-four participants ($18.6\%$) answered affirmatively. To ensure that these suspicions did not drive the observed lack of treatment effects, we conducted a robustness check excluding these participants. As detailed in Section 3.2, excluding these participants did not alter the substantive findings. Sociodemographic characteristics of the final sample are reported in Appendix 2. Consistent with our preregistered analysis plan, we retained the full sample for the primary analysis to avoid post hoc selection bias and preserve randomization. Participants were distributed near-equally across the human (n = 114) and AI (n = 122) treatment conditions and across the three bot strategies within each label. The sociodemographic profile of the sample was broadly balanced: the modal age groups were 28–37 (41 \%) and 18–27 (29 \%); sex was evenly split; and the sample showed substantial ethnic and national diversity. Most participants were employed full-time or part-time, with roughly one-third identifying as students.

%\begin{table}[htbp]
%\centering
%\caption{Distribution of participants by treatment and bot strategy\label{tab:treat-bot}}
%\begin{tabular}{lll}
%\toprule
%\textbf{Treatment} & \textbf{Bot strategy} & \textbf{n} \\
%\midrule
%AI     & Unconditional cooperator & 41 \\
%       & Conditional cooperator   & 42 \\
%       & Free-rider               & 39 \\
%\addlinespace
%human  & Unconditional cooperator & 38 \\
%       & Conditional cooperator   & 35 \\
%       & Free-rider               & 41 \\
%\bottomrule
%\end{tabular}
%\end{table}

\subsection{Preregistration and Sample Deviations}
The study design, hypotheses, and analysis plan were preregistered on AsPredicted (registration \#234846). The study design and hypotheses remained identical to those specified in the preregistration. Regarding sample size, the preregistration stated that it would be determined by an a priori power analysis based on pilot effect sizes. However, pilot data revealed effect sizes close to zero. Because standard power analyses for near-zero effects yield unrealistically large sample requirements, we could not proceed with the pre-specified power calculation. Instead, we targeted a sample size sufficient to detect substantively meaningful deviations in cooperation. The final sample of 236 complete group interactions yields approximately 40 independent observations per cell for the main strategy comparisons. Post-hoc sensitivity analysis confirms that this yields high precision for our primary investigation: the 90\% confidence interval for the label effect allows us to rule out label effects larger than approximately 4 tokens. While the design is less powered to detect subtle interaction effects among bot strategies, the confidence intervals are sufficiently narrow to exclude large, disruptive behavioural shifts. In addition, the equivalence test reported in Section 3.2 was not preregistered: we specified the smallest effect size of interest as ±5 tokens (5\% of the endowment, corresponding to approximately 0.24 standard deviations of participants' mean contributions) and conducted two one-sided tests (TOST) against these bounds. All deviations from the preregistration are fully reported here, and the complete preregistration document is available at [https://aspredicted.org/wd7j-jyg5.pdf].

\subsection{Measures}
% Dependent Variables
The primary dependent variables were derived from participants’ decisions in the PGG and the Prisoner’s Dilemma (PD). In the PGG, participants made contribution decisions ranging from 0 to 100 tokens over ten rounds. These contributions were analysed both as repeated measures in mixed-effects regressions and as mean contributions across rounds in linear regressions. The PD provided a binary outcome as either cooperate or defect, which was coded accordingly and analysed using logistic regression models.
% Norm Elicitation
Beyond behavioural outcomes, we also measured participants’ normative perceptions following the two decision tasks. Building on \cite{krupkaIdentifyingSocialNorms2013}, participants rated the social appropriateness of five possible contribution levels (0, 25, 50, 75, and 100 tokens) not specific to human or AI agent contributions, on a Likert scale ranging from very socially inappropriate to very socially appropriate. These ratings were used to compute an overall norm score and a norm slope, indicating the extent to which perceived appropriateness increased with contribution size. We additionally measured two complementary norm expectations: empirical expectations, capturing participants’ beliefs about others’ actual contributions, and injunctive expectations, capturing their beliefs about how much others thought one should contribute. Together, these three measures provided quantitative indicators of normative expectations and their relationship to both treatment conditions and cooperative behaviour.

% Survey Measures
Following the behavioural tasks, participants completed a post-experiment survey assessing perceived trust, fairness, group cohesion, and normative pressure during the game. Participants in the AI treatment received additional items on the AI player’s social perception, accountability, representation, and fairness. Each construct was measured with two to three items and aggregated into index scores. The full set of study instruments and constructs is summarized in Table \ref{tab:instruments}. These exploratory measures were designed to identify potential mechanisms underlying variation in cooperation and norm perception across conditions.
\bigskip

\section{Results}
\subsection{Experiment Descriptive Results}
\begin{figure}[H]
  \centering
  \includegraphics[width=1\textwidth]{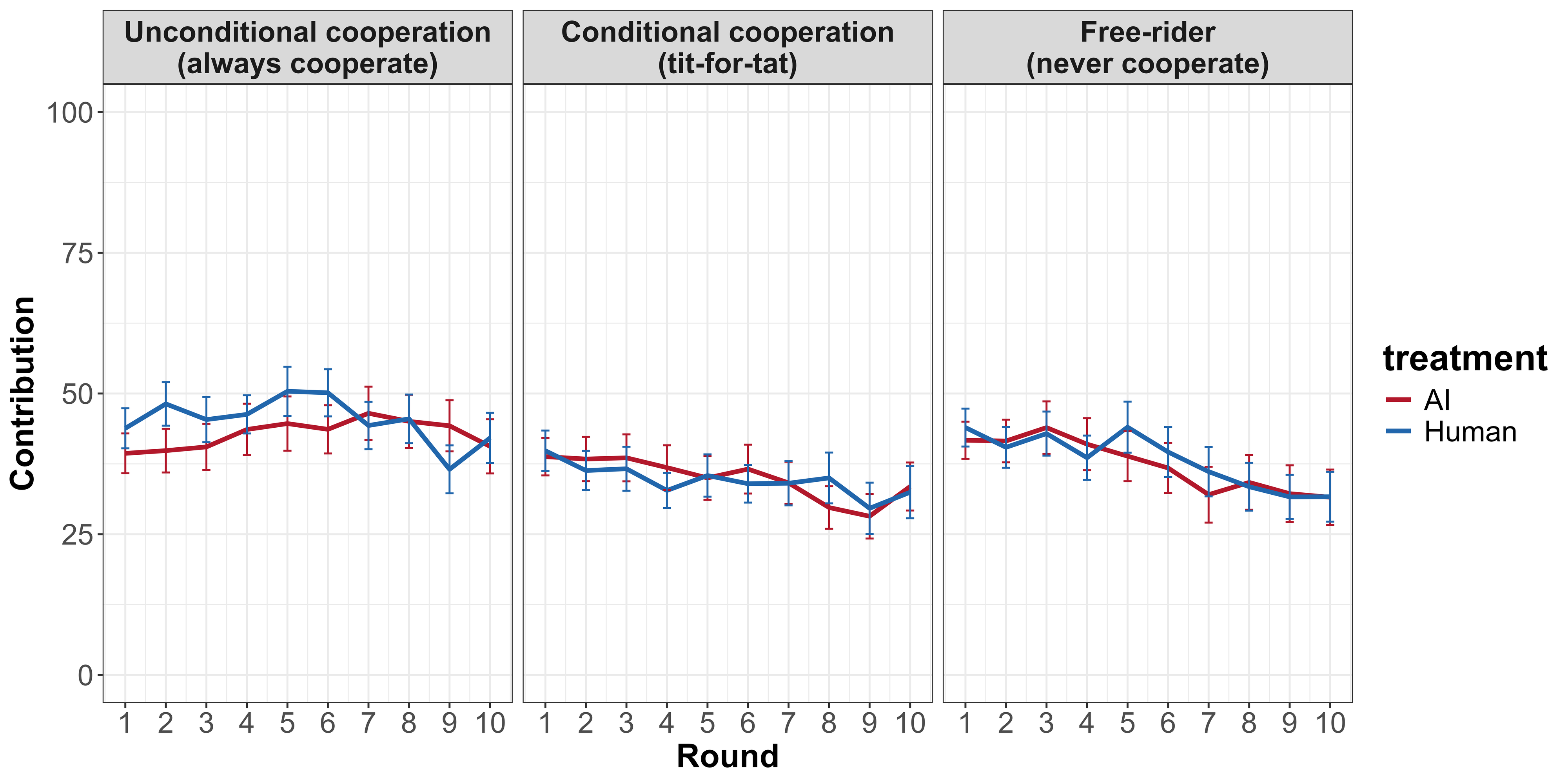}
  \caption{Mean Contribution per Round by Treatment and Bot Strategy.}
  \label{fig:PGG_rounds}
\end{figure}

% Descriptive PGG 
Figure \ref{fig:PGG_rounds} shows that average contributions in the PGG were very similar across human and AI groups. Across all conditions, contributions started at roughly 40–50 tokens and declined modestly over the ten rounds to around 30–40 tokens, consistent with the typical downward trend in repeated public-goods interactions. Groups paired with the unconditional cooperator bot tended to sit at the upper end of this narrow range, conditional cooperators at the lower end, and free-riders in between, but the trajectories largely overlapped and followed the same gradual decline. Human-labelled groups contributed slightly more than AI-labelled groups, yet these gaps were only a few tokens.

\begin{figure}[H]
  \centering
  \includegraphics[width=1\textwidth]{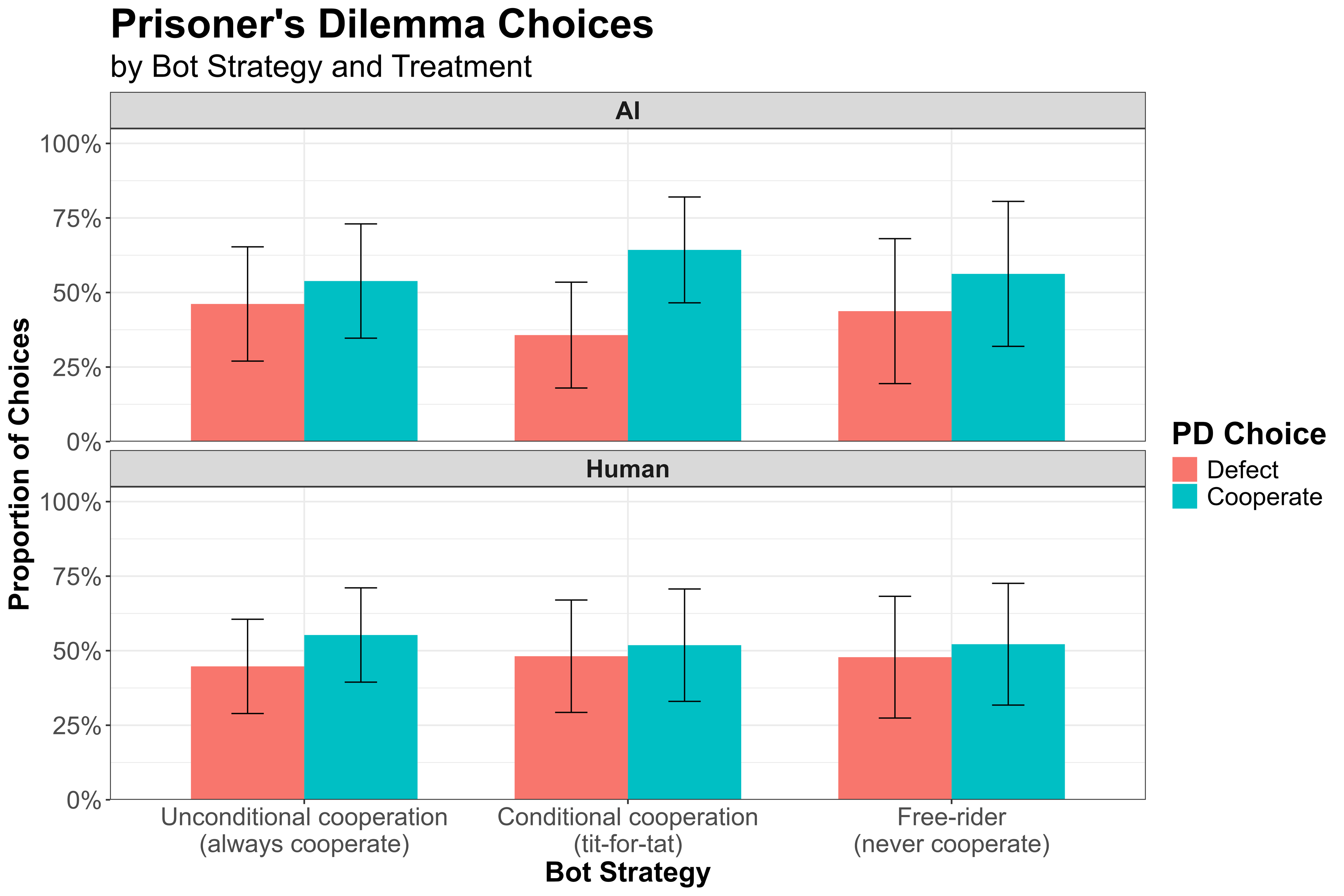}
  \caption{Prisoner's Dilemma choices by treatment and bot strategy.}
  \label{fig:PD_Choices}
\end{figure}

% PD Results 
Figure \ref{fig:PD_Choices} shows cooperation and defection rates in the one-shot Prisoner’s Dilemma. Across both human and AI treatments, cooperation was slightly more common than defection, and overall rates were similar. In the AI–conditional cooperation condition, about two-thirds of participants cooperated, whereas in the human treatment cooperation rates clustered around 50\% across strategies. 

\begin{figure}[H]
  \centering
  \includegraphics[width=1\textwidth]{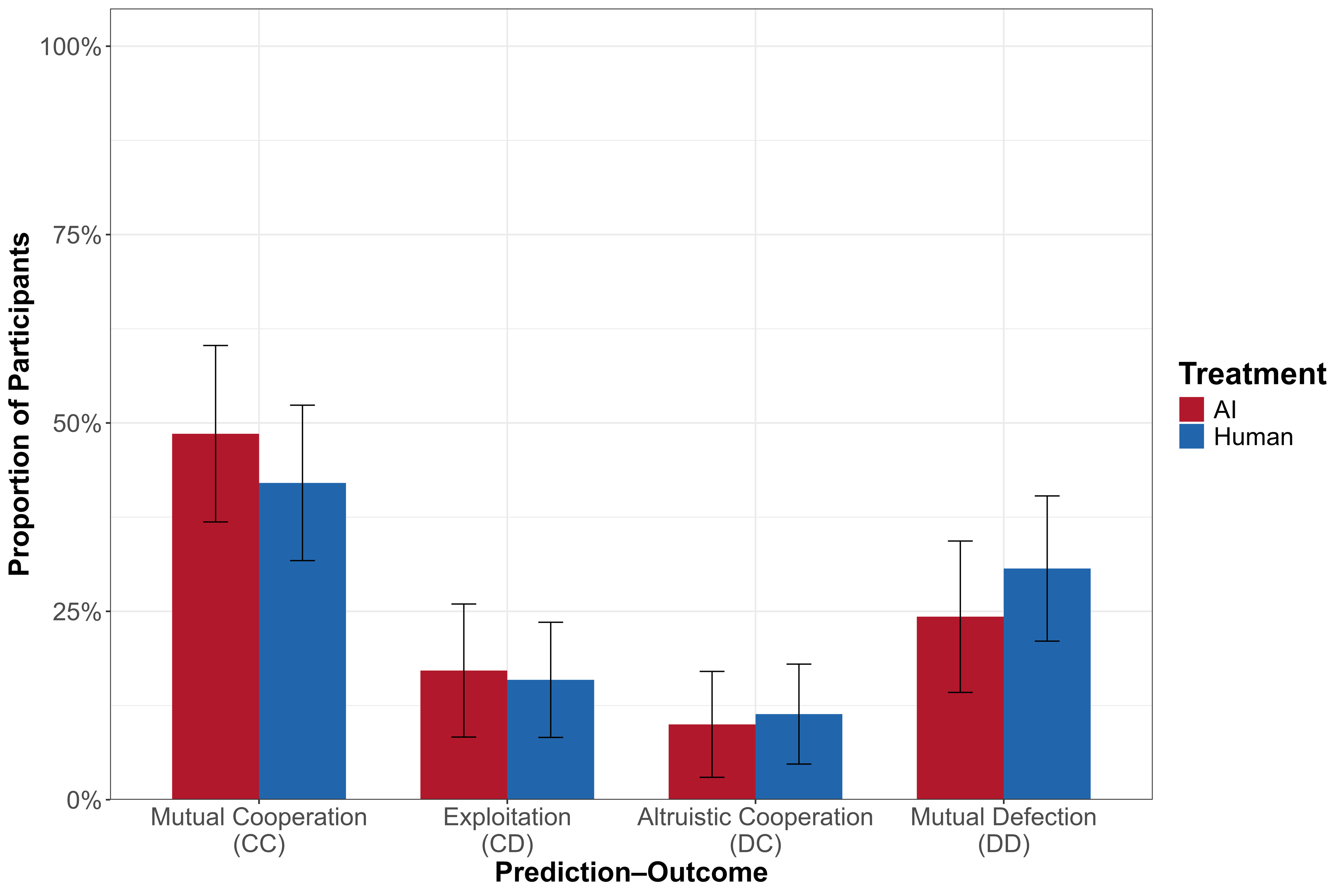}
  \caption{Prisoner's Dilemma choices by predicted partner behaviour and own decision. The first letter denotes the prediction of the partner's behaviour, and the second letter denotes the participant's choice.}
  \label{fig:PD_pred}
\end{figure}

% PD Prediction Results
Figure \ref{fig:PD_pred} shows the distribution of participants' Prisoner's Dilemma decisions by the expectation of their partner's choice. The most common pattern was CC (mutual cooperation), accounting for nearly half of all cases in both treatments, although it was slightly more frequent in the AI condition. On the other hand, around a quarter of participants fell into the category of DD (mutual defection), expecting defection and defecting themselves, which is somewhat more frequent in the human condition. Less common were CD (exploitation), in which participants predicted their partner's cooperation but defected, and DC (altruistic cooperation), in which participants predicted defection but cooperated anyway. Both occurred relatively rarely at around 10-20\%. Overall, the distribution of the profiles suggests that participants were more inclined towards mutuality, either cooperating or defecting, compared to exploitation or altruism. 

\subsection{Model-based Results}
\subsubsection{PGG}
To examine whether the presence of an AI-labelled agent changed cooperation decisions, we estimated a series of statistical models across the PGG, the Prisoner’s Dilemma, and the norm expectations elicited from participants. The models allow us to disentangle how perceptions of agent type and bot strategies interact with normative pressures on cooperation decisions in a group (PGG), how cooperation persists in subsequent interactions (PD), and how normative expectations and perceptions might mediate these dynamics. 

% PGG Full Model
First, we examine the main effect of our treatments on cooperation decisions using a linear mixed-effects regression model. To this end, we include the treatment and bot strategies, others' contributions in previous rounds (group pressure), their own contributions in the previous round (individual inertia), and time trends across rounds. Because each participant made repeated decisions across rounds, observations were nested within individuals. We therefore used mixed-effects models with participant-level random intercepts and random slopes for others’ previous contributions and round number. This approach captures both baseline differences in contribution levels and individual variation in responsiveness to prior group contributions and time trends. 

Our primary hypothesis (H1) predicted a "differentiation effect," where the presence of an AI-labelled agent would reduce overall cooperation and normative pressure. Contrary to this prediction, the mixed-effects model (see Table \ref{tab:mfinal}) reveals no significant effect of the AI label on contribution levels (human vs AI; $β = 1.09, p = .738$). Consequently, we reject H1; we find no evidence that the AI label reduced contribution levels. The regression further shows that participants' behaviour was strongly shaped by both their own and others' prior aggregate contributions. They contributed significantly more when others had contributed highly in the previous round ($β = 0.14, p = .006$) and when they had contributed more in the previous round themselves ($β = 0.28, \textless p .001$). We also observe a round effect, with the typical decline in cooperation in the repeated game ($β = –0.59, p = .001$). 

We further hypothesised that the specific strategy of the agent would shape group norms: unconditional cooperation would raise contributions (H1a), conditional cooperation would sustain them (H1b), and free-riding would erode them (H1c).
The results offer little support for these specific predictions. While groups with unconditional cooperators trended slightly higher and free-riders slightly lower (see Figure \ref{fig:PGG_rounds}), these differences were not statistically significant in the regression model. The lack of substantial differentiation among bot strategies may reflect both the presence of two other human group members and the aggregate-feedback structure, which diluted the behavioural signal of the single automated agent. This finding is robust to participant suspicion: Excluding the 44 participants who expressed doubt about the group composition, the treatment effect remained statistically non-significant ($\beta = 3.05, p = .396$) and the primary behavioural drivers (inertia, conditional cooperation) remained stable. These results indicate that we find the same behavioural rules which govern conditional cooperation, namely sensitivity to others’ contributions, individual inertia, and decline across rounds, equally in both human and AI groups.

\begin{table}[H]
\centering
\caption{Public Goods Game Contributions (Linear Mixed-Effects Regression Model)}
\label{tab:mfinal}
\begin{tabular}{lccc}
\toprule
\textbf{Predictors} & \textbf{Estimates} & \textbf{CI} & \textbf{$p$} \\
\midrule
\multicolumn{4}{l}{\textit{Fixed effects}} \\
(Intercept)                                         & 25.07 & 18.06 -- 32.08 & \textbf{\textless 0.001} \\
Others' lagged contrib.                             &  0.14 &  0.04 --  0.24 & \textbf{0.006} \\
human (vs. AI)                                      &  1.09 & -5.31 --  7.49 & 0.738 \\
Conditional Cooperation (vs. Unconditional)         & -2.82 & -8.24 --  2.59 & 0.307 \\
Free-Rider (vs. Unconditional)                      &  1.10 & -4.54 --  6.73 & 0.703 \\
Own lagged contrib.                                 &  0.28 &  0.23 --  0.32 & \textbf{\textless 0.001} \\
Round (centered)                                    & -0.59 & -0.92 -- -0.26 & \textbf{0.001} \\
human $\times$ Others' lagged contrib.              & -0.02 & -0.14 --  0.11 & 0.803 \\
\midrule
\multicolumn{4}{l}{\textit{Random Effects (variance)}} \\
$\sigma^2$ (Residual)                   & \multicolumn{2}{c}{272.51} & \\
$\tau_{00}$ id                          & \multicolumn{2}{c}{139.48} & \\
$\tau_{11}$ id.Others lagged contrib.   & \multicolumn{2}{c}{0.02}   & \\
$\tau_{11}$ id.Round (centered)         & \multicolumn{2}{c}{1.99}   & \\
\midrule
ICC                                     & \multicolumn{2}{c}{0.34}   & \\
$N_{\text{id}}$                         & \multicolumn{2}{c}{236}    & \\
Observations                            & \multicolumn{2}{c}{2,124}  & \\
Marginal $R^2$ / Conditional $R^2$      & \multicolumn{2}{c}{0.165 / 0.448} & \\
\bottomrule
\end{tabular}

\vspace{4pt}
{\footnotesize
\textit{Notes.} Linear mixed model (REML) with random intercepts and random slopes for \emph{others} and \emph{round} by participant.
Confidence intervals are 95\%. Reference categories: Partner label = \emph{AI}; Bot strategy = \emph{Always coop}.
Model: \texttt{player $\sim$ others * treatment + bot\_strategy + lagp + round + (1\,|\,unique\_id) + (0 + others,|\,unique\_id) + (0 + round,|\,unique\_id)}.
}
\end{table}

\paragraph{Precision and Equivalence of Treatment Effects:}
% Null results
While the mixed-effects model revealed no significant treatment effect ($β = 1.09, p = .738$), a non-significant difference does not by itself demonstrate equivalence. We therefore formally tested whether contributions in the two label conditions were equivalent within a smallest effect size of interest of ±5 tokens (5\% of the endowment; see Section 2.4) using two one-sided tests (TOST) on the estimated marginal means contrast. The contrast between conditions was small (AI – human = –0.40, SE = 2.03, 90\% CI [–3.75, 2.95]), and both one-sided tests rejected differences at the equivalence bounds ($t(232.3) = 2.27$, $p = .012$ against –5; $t(232.3) = –2.66$, $p = .004$ against +5): contributions were statistically equivalent within ±5 tokens in the full sample. Participants, therefore, appeared to approach the cooperation task under shared normative expectations, regardless of whether one group member was labelled as human or AI.

\subsubsection{Prisoner's Dilemma}
\begin{figure}[H]
  \centering
  \begin{subfigure}{0.48\textwidth}
    \includegraphics[width=\linewidth]{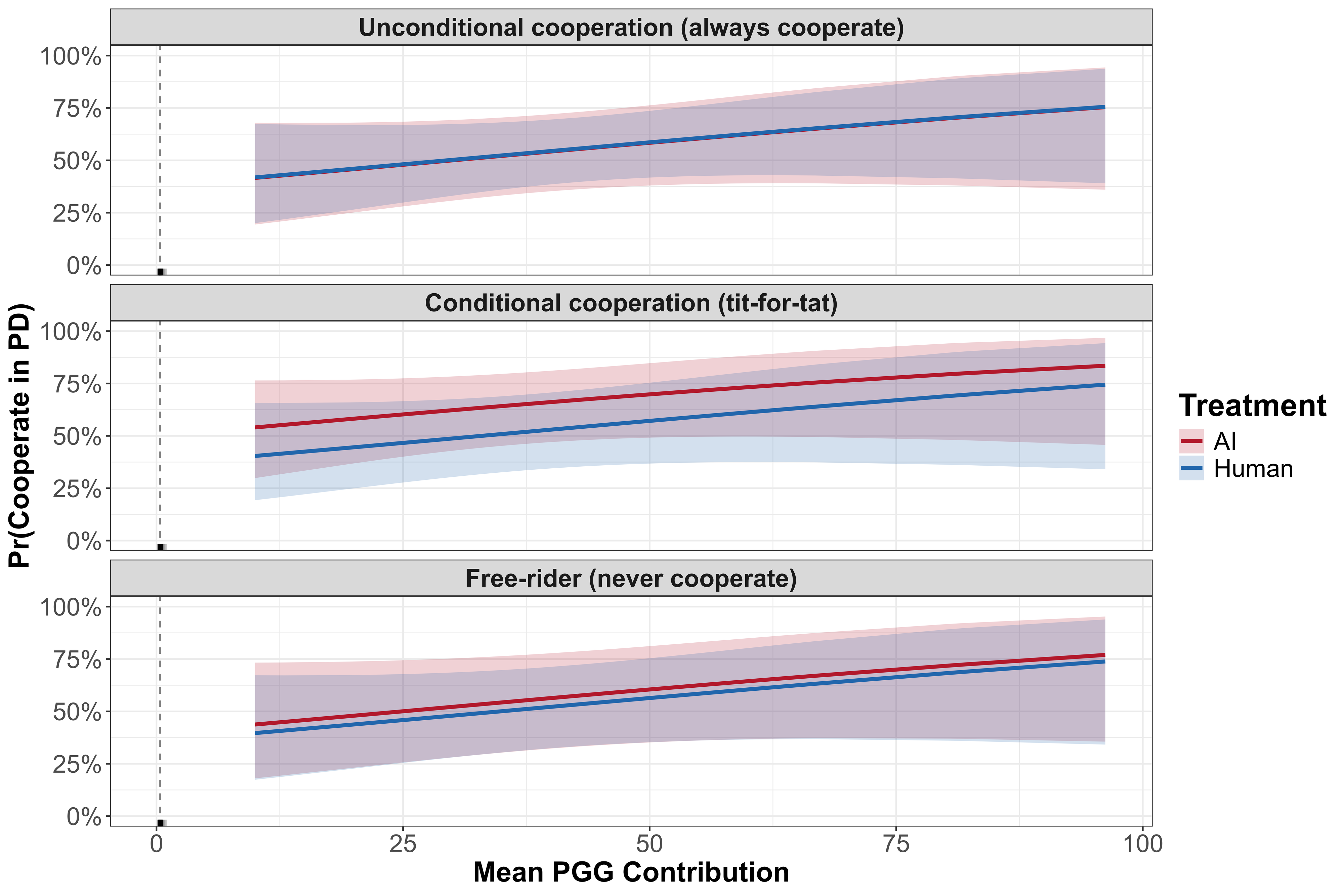}
    \caption{Individual contribution}
    \label{fig:PD_indi}
  \end{subfigure}
  \hfill
  \begin{subfigure}{0.48\textwidth}
    \includegraphics[width=\linewidth]{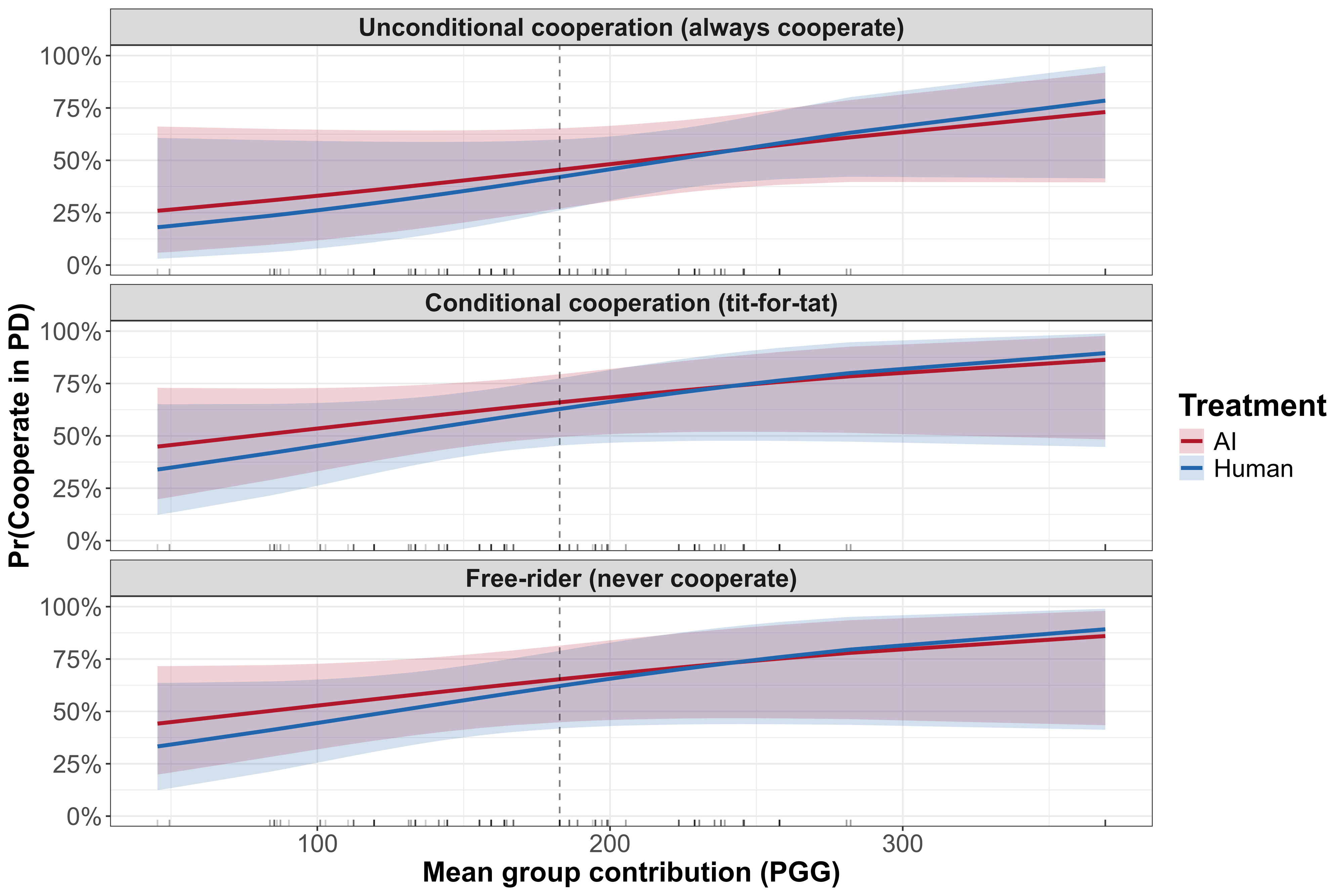}
    \caption{Group contribution}
    \label{fig:PD_grp}
  \end{subfigure}
  \caption{Predicted probability of PD cooperation by prior PGG contributions at the (a) individual and (b) group level.}
  \label{fig:PD_combined}
\end{figure}

% PD Results
Hypothesis H2 predicted that cooperative norms formed in human-only groups would be more robust, leading to higher norm persistence (cooperation in the one-shot PD) compared to mixed groups. However, the logistic regression analysis contradicts this prediction. As illustrated in Figures \ref{fig:PD_indi} and \ref{fig:PD_grp}, while the probability of cooperating in the PD increases with prior contributions at both the individual (Panel a) and group levels (Panel b), the regression lines for human and AI treatments overlap almost perfectly. This visual convergence confirms that the previous group composition did not significantly predict the likelihood of cooperating. Therefore, the persistence of the group experience's norm did not differ by agent type. Consequently, H2 is not supported; the normative inertia carried over into the subsequent interaction regardless of whether the previous group included an AI.

\subsection{Complementary Analyses}
\subsubsection{Norm Attitudes}
% Norm Expectations & Perceptions
We asked participants a few questions about perceptions of cooperative norms. After the two games, they were asked to rate different contribution levels (0, 25, 50, 75, 100) and the social acceptability of those contributions. Further, they were asked what they believed other people in their group contributed (empirical expectation) and what they thought they were expected to contribute (normative expectation), allowing us to test normative pressures across conditions. 

\begin{figure}[H]
  \centering
  \includegraphics[width=1\textwidth]{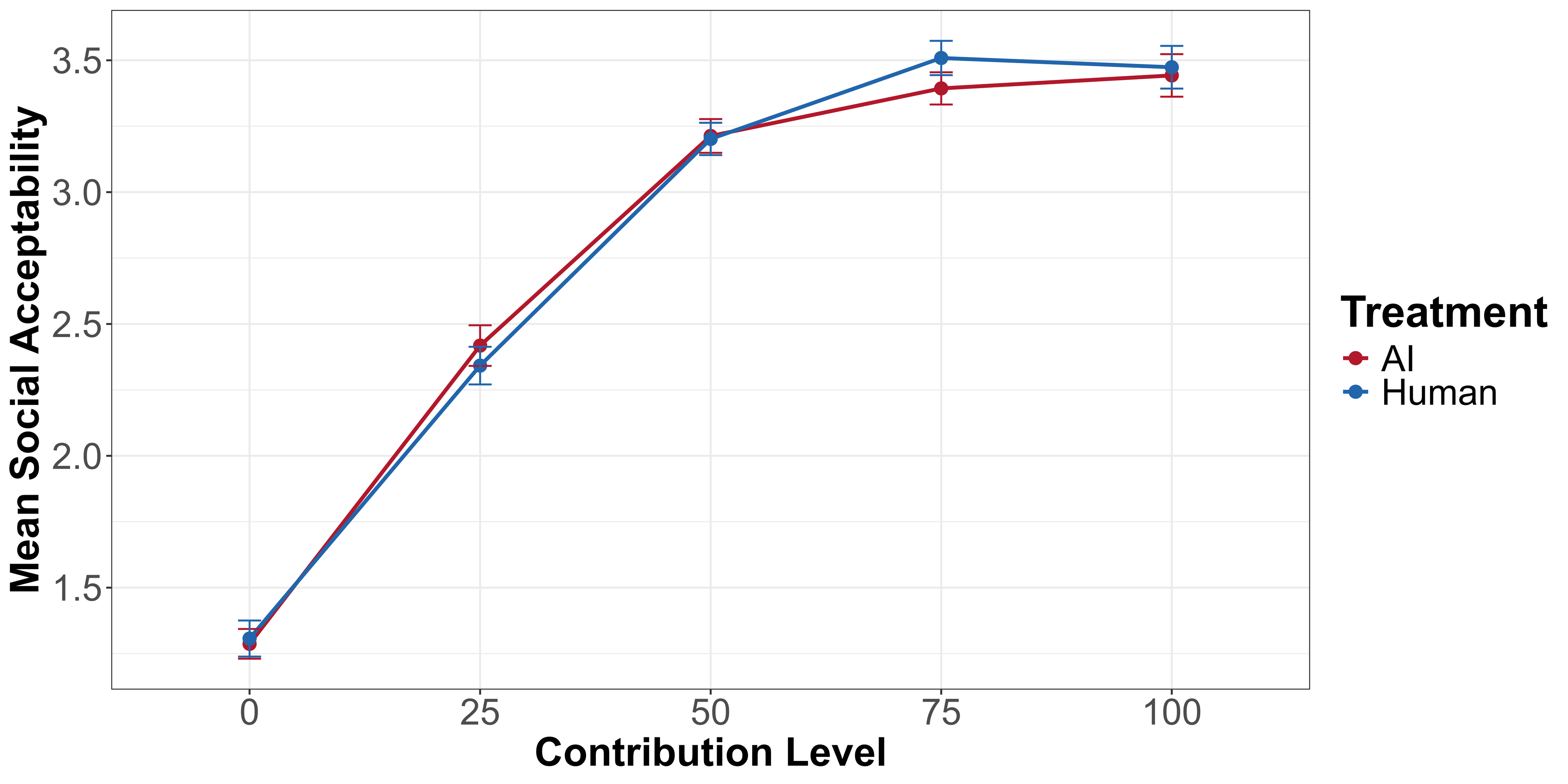}
  \caption{Social acceptability of different contribution levels by treatment}
  \label{fig:SocAcc}
\end{figure}

%% Social Acceptability
Participants’ social acceptability ratings closely aligned with contribution levels (see Figure \ref{fig:SocAcc}). Contributions of 0 were judged as very unacceptable (M = 1.3), whereas acceptability rose sharply between 25 (M = 2.38) and 50 (M = 3.21). We then see a plateauing of ratings with almost identical ratings for contributions of 75 and 100 (M = 3.45). A regression analysis found no statistically significant differences between the human and AI groups, with acceptability ratings closely aligned and diverging only slightly at the 75 contribution level.

\begin{figure}[H]
  \centering
  \includegraphics[width=1\textwidth]{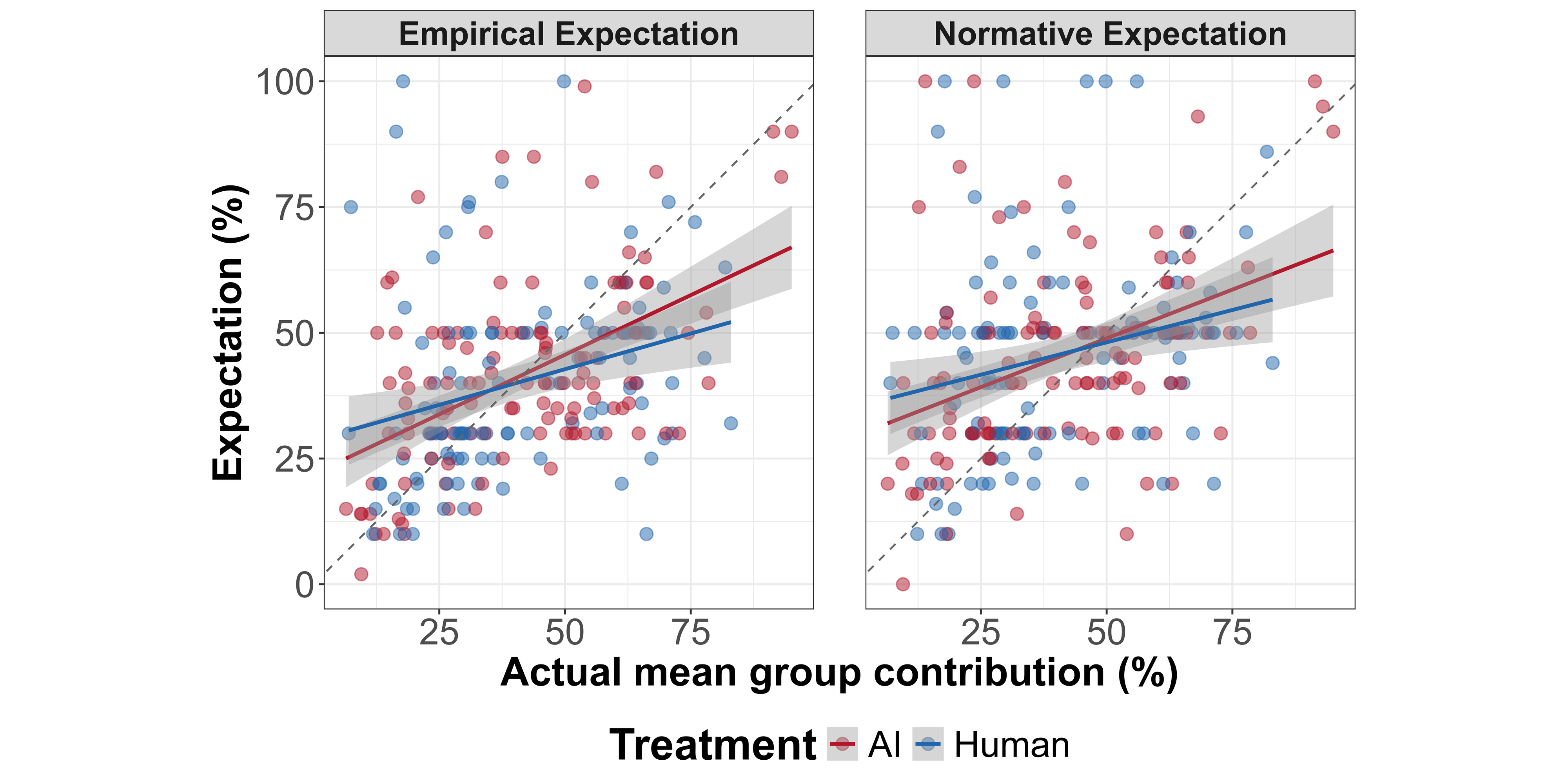}
  \caption{Norm Expectations by Treatment}
  \label{fig:NormExp}
\end{figure}

%% Empirical Expectations
Participants’ empirical expectations closely matched actual group contributions (Median Expectation = 40.8, Median Actual = 41.0), indicating overall accurate beliefs about others’ cooperation. Differences between treatments were negligible: participants slightly overestimated others in the AI condition (+0.3) and slightly underestimated them in the human condition (–0.8). As shown in Figure \ref{fig:NormExp}, expectations rose linearly with actual contributions, with nearly identical slopes across treatments.

%% Normative Expectations
Participants’ injunctive norm expectations exceeded the group's actual contributions (Median Expectation = 45.6, Median Actual = 41.0), indicating that they believed others should contribute slightly more than they did. Figure \ref{fig:NormExp} shows that this pattern was consistent across treatments: both AI (Δ = +4.3) and human (Δ = +4.8) groups showed similar positive gaps, with largely overlapping regression lines. Overall, participants’ normative beliefs aligned with actual cooperation levels but reflected modestly higher expectations regarding contributions. Since normative expectation patterns mirrored actual contributions and were nearly identical across treatments, this further supports the idea that the group's normative environment operated similarly regardless of the AI label.

\subsubsection{Post-Experiment Survey}
\begin{figure}[H]
  \centering
  \includegraphics[width=1\textwidth]{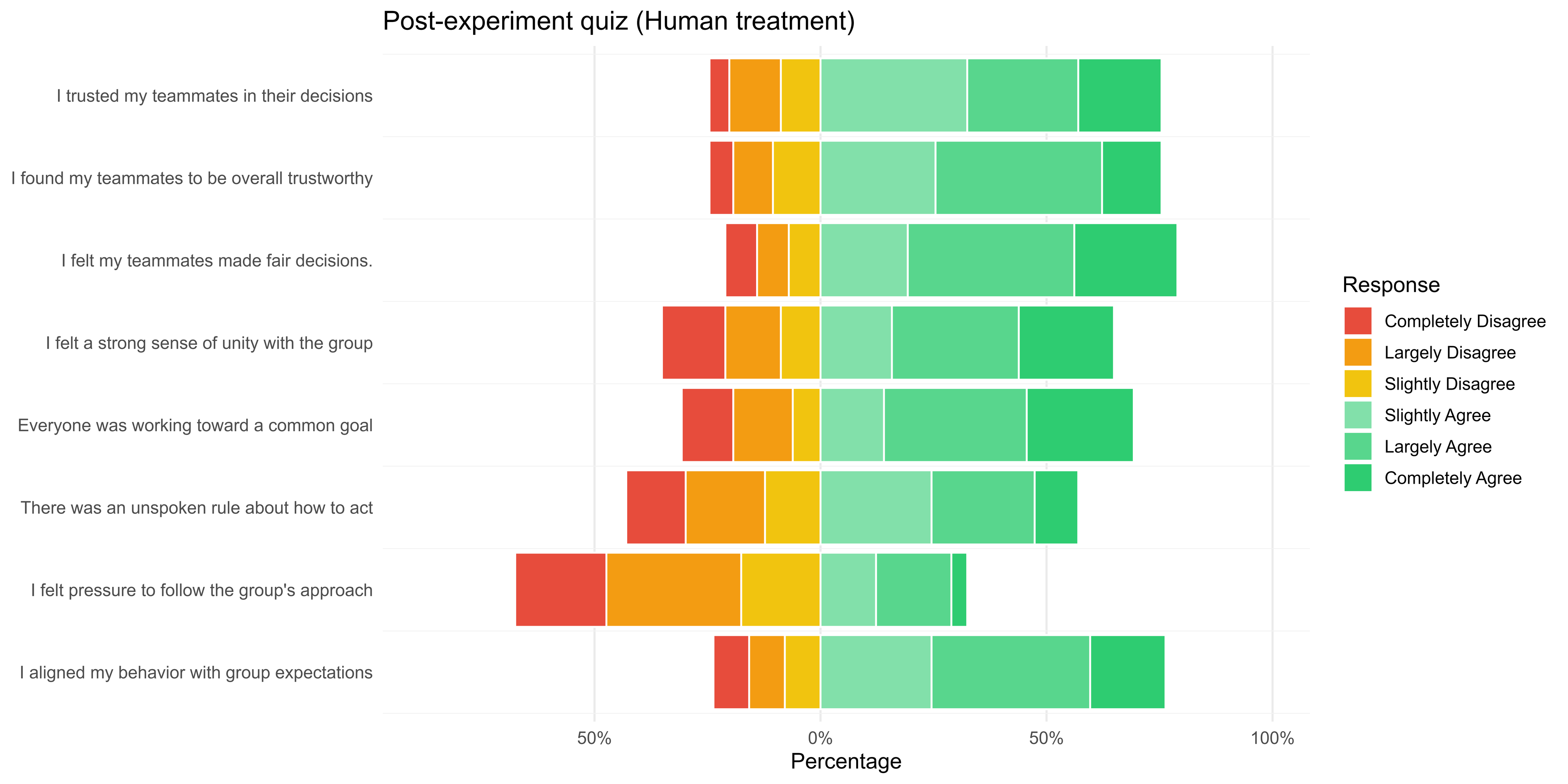}
  \caption{Post Experiment Survey: Human Treatment}
  \label{fig:PQhum}
\end{figure}

\begin{figure}[H]
  \centering
  \includegraphics[width=1\textwidth]{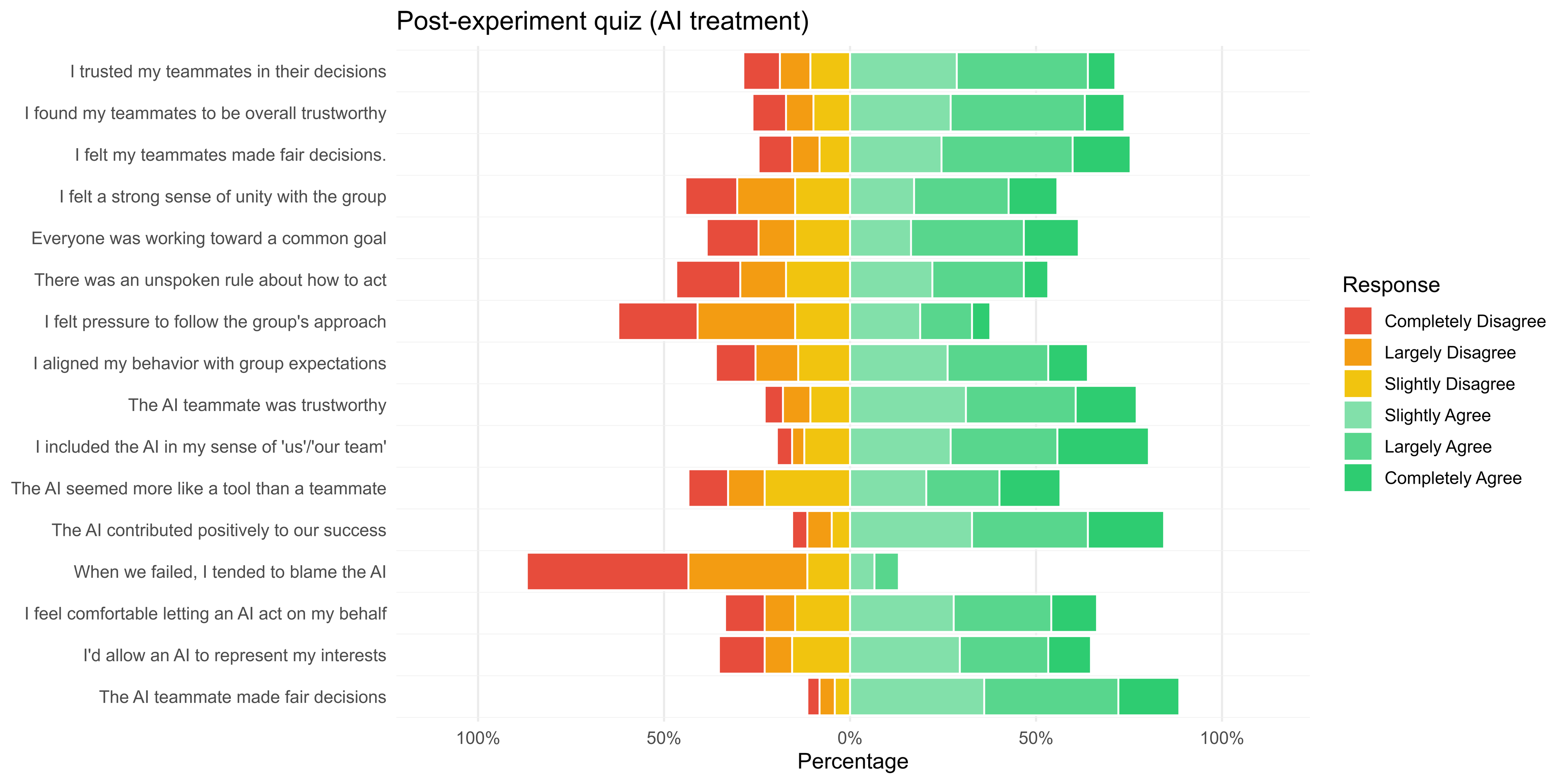}
  \caption{Post Experiment Survey: AI Treatment}
  \label{fig:PQAI}
\end{figure}

% Post-Experiment Survey
Several indicators focused on trust, perceptions and acceptability were elicited from participants to gain a deeper understanding of their motivations within the games. The results indicate positive group perceptions across both treatments (Figure \ref{fig:PQhum} \& Figure \ref{fig:PQAI}). Participants in both treatments agreed that their teammates were fair, trustworthy and cooperative, while responses were more mixed regarding normative pressures. Mean responses for trust, fairness, cohesion, and normative pressure items did not differ significantly across treatments (all p \textgreater  .10), except for the statement “I aligned my behaviour with what I thought the rest of the group expected of me”. Here we find a marginally lower response in the AI condition ($β = –2.17, p = .031$). This suggests that participants interacting with an AI-labelled teammate felt slightly less alignment pressure, while general perceptions of fairness and trust remained comparable across treatments. The participants in the AI treatment found AI to be trustworthy, fair, and part of the team, although its role was seen more as a tool than as a teammate. Interestingly, most participants thought that AI contributed positively to the group's success and wouldn't blame AI for failures. 

To test whether any of these attitudes and perceptions influenced cooperative behaviour in the PGG, we estimated two regression models. For the general statements asked of both treatments, we found that trust was the only significant predictor of higher contributions ($β = 3.71, p = .046$). Among the AI treatment variables, we found that cooperation increased with trust ($β = 7.33, p = .006$) and normative pressure ($β = 4.86, p = .008$). Yet we found that overall AI acceptance would be associated with lower contribution levels ($β = –5.76, p = .001$). Together, these findings suggest that participants who trusted their group and perceived stronger normative expectations contributed more. In contrast, in the AI group, we found that algorithm aversion can adversely affect contributions. 

% \subsection{Exploratory Analysis}
%Things I have have looked at which did not yield any differences:\\
%- \textbf{AI Supporter vs AI Skeptic}: Do AI attitudes categorised into two categories (below/above median) influence how much people contribute in interaction with the others contribution\newline
%- \textbf{Social perception of AI}: Does social perception of AI influence contributions and norm alignment to others contribution. \newline
%- \textbf{AI Delegation or diffusion of responsibility}: Does trusting AI mean they "delegate to AI" and don't contribute themselves? \newline
%- \textbf{Convergence/Polarisation}: Is there a difference between human and AI when it comes to converging on contributions? \newline

\bigskip

\section{Discussion}
Our study examined whether the inclusion of an AI-labelled teammate alters the social dynamics of group cooperation. Contrary to H1, which was derived from prior work on algorithm aversion and reduced cooperation with AI partners \citep{dietvorstAlgorithmAversionPeople2015, karpusAlgorithmExploitationHumans2021, bazaziAIsAssignedGender2025}, we found no systematic behavioural differences between human- and AI-labelled conditions. Participants’ contributions followed familiar normative patterns such as conditional cooperation, sensitivity to others’ past behaviour, and gradual decline over time. These observed patterns were statistically indistinguishable across treatments and strategies. This pattern mirrors findings from prior dyadic studies \citep{ngWhenCommunicativeAIs2023, makoviTrustHumanmachineCollectives2023} showing that trust and reciprocity can generalize to AI partners. When the AI was embedded in a collective context with anonymous aggregate feedback, its label had no detectable influence on cooperative behaviour: participants appeared to respond to the group's behaviour rather than to the agent's identity.

\subsection{Bounded Normative Equivalence}

These results point to a limited form of bounded normative equivalence. By this, we mean that the observed cooperation patterns, norm perceptions, and norm persistence measures were highly similar across human- and AI-labelled conditions, with contributions formally equivalent within ±5 tokens in the full sample. We introduce the term 'normative equivalence' to describe the observed process-level similarity in how social norms guide cooperation behaviour. By bounded normative equivalence, we do not argue that humans perceive AI as morally equivalent to, equally trustworthy as, or socially interchangeable with human partners. Rather, the concept refers explicitly to a form of norm adherence, the mechanism by which individuals align their cooperation with empirical and injunctive expectations \citep{bicchieriSocialNorms2018}. It denotes that the same behavioural regularities of reciprocity and conditional cooperation emerged regardless of the partner's label. We distinguish this from norm enforcement. Since our design excluded peer punishment, our findings indicate that humans complied with cooperative norms similarly in the mixed groups studied here. However, enforcement mechanisms appear necessary when the human social buffer is removed. \cite{makoviRewardsPunishmentsHelp2025} demonstrates that while punishment alone increases cooperation with AI, it does not eliminate the machine penalty; only the combination of peer rewards and punishment successfully closes the gap. This suggests that while bounded normative equivalence emerged without explicit enforcement in our mixed groups, achieving equivalence in cooperation levels within AI-dominated contexts may require explicit, combined enforcement mechanisms.

These findings bound the scope of the 'differentiation' perspective often found in human-AI interaction research, particularly assumptions regarding algorithm aversion and moral disengagement \citep{karpusAlgorithmExploitationHumans2021, mutznerEvadingAlgorithmIncreased2023}: such differentiation may apply only in specific contexts and was not evident in our group setting with anonymous aggregate feedback. This aligns with recent experimental work using the Prisoner's Dilemma, which similarly found no significant differences in cooperation rates between human and AI partners \citep{bazaziAIsAssignedGender2025}. However, this equivalence may depend on the group's social density: while \cite{makoviRewardsPunishmentsHelp2025} observed a 'machine penalty' in groups where participants believed all partners were machines, our results suggest that in mixed groups where humans remain the majority, the presence of human peers buffers against this effect. Contrary to predictions derived from Social Identity Theory or Mind Perception \citep{oudahPerceptionExperienceInfluences2024}, which suggest reduced obligation toward non-human agent participants, we found no evidence that the AI label reduced aggregate cooperation, trust-related responses, or norm alignment. This suggests that bounded normative equivalence is distinct from the surface-level social responses described by the Computers Are Social Actors (CASA) paradigm \citep{nassComputersAreSocial1994}. While CASA focuses on unconscious reactions to anthropomorphic cues, bounded normative equivalence denotes a deeper behavioural alignment in which cooperation follows the group's functional logic (reciprocity) rather than the partner's ontological category. Therefore, in the repeated-group setting with aggregate feedback studied here, the 'AI' label made no detectable behavioural difference. Within limited information environments, participants appear to rely on group-level observable behaviour as the primary normative cue, without distinguishing among individual team members.

\subsection{Theoretical and Methodological Implications}

The informational structure of the setting is central to interpreting these results: rather than a mere methodological detail, it constitutes a substantive boundary condition of the normative equivalence we observed. Because participants received feedback on the group's total contribution rather than on individual actions, no member's individual behaviour could be identified (attribution ambiguity), and the single agent's strategy was blended with that of two human members (signal dilution). Under these conditions, cooperation appeared to be guided by dynamic collective expectations rather than by isolated actions or categorical distinctions. Many real-world collective settings share this opacity: in online communities, markets, and large collaborative platforms, individual contributions are often not identifiable. The experiment thus provides a baseline in which cooperative dynamics remained normatively equivalent across human and AI labels under conditions of minimal social presence, no communication, and anonymous aggregate feedback. However, these dynamics may change as the group's composition shifts. For instance, \cite{makoviRewardsPunishmentsHelp2025} observe a clear 'machine penalty' when participants believe all partners are machines, suggesting that the presence of a human majority in our study may have buffered against such effects. Increasing the proportion of AI agents could therefore alter perceived social balance, the diffusion of responsibility, or majority influence, potentially amplifying or diminishing normative pressures. Future studies can build on this baseline by introducing adaptive, communicative, or emotionally expressive AI agents and by systematically varying their proportions within groups. This would allow testing of when, and through which mechanisms, normative equivalence might begin to break down.

From a broader perspective, these findings speak to the integration of AI into human collectives. It further fits into the emerging field of Machine Behaviour \citep{rahwanMachineBehaviour2019,tsvetkovaNewSociologyHumans2024}, illustrating how artificial agents may be functionally integrated into human collectives without requiring complex social intelligence. The stability of the hybrid groups suggests that 'socialness' in a system is not solely a property of agents' minds but an emergent property of the rules and feedback loops governing their interactions. Social norms governing cooperation might therefore be more elastic than might initially be assumed. In our setting, individuals readily applied the same cooperative logic to heterogeneous groups that include artificial actors. This elasticity may prove beneficial as AI systems become routine participants in work teams and online communities and are more involved in decision-making processes. Yet it also raises new questions about accountability and transparency. If cooperative norms extend to AI systems as readily as they did under the conditions studied here, it should be considered whether responsibility for outcomes may diffuse just as easily among human and non-human participants.
\bigskip

\subsection{Limitations}
Based on our experiments and results, several limitations should be acknowledged. First, as with many online experiments involving deception, there remains a risk that some participants did not fully believe in the group composition. Although most participants correctly identified their condition, a minority ($18.6\%$) expressed doubt about the group's composition. However, our robustness check excluding these participants indicated that the main findings remained substantively unchanged: the treatment contrast remained non-significant and directionally similar, although it was estimated with less precision in the smaller subsample. Yet, this highlights the inherent difficulty of creating credible mixed-agent group settings in online environments, where subtle cues of artificiality or repetition can influence perceived realism.
Second, our design focused on short-term, anonymous interactions. Without extended histories or reputation-building, participants’ behaviour may have reflected situational cooperation rather than deeper internalization of norms. Real-world human–AI collaboration can involve ongoing relationships and feedback which cannot be fully captured in brief experimental sessions.
Third, while our treatments varied both the agent label and strategy, the AI’s behaviour was scripted rather than adaptive. This limits the ecological validity of our findings, as real AI systems increasingly learn and respond dynamically to human input. Future studies could incorporate adaptive agents to examine whether evolving responsiveness strengthens or weakens normative alignment over time. Specifically, adaptations involving communication or punishment mechanisms that heighten the perception of complex interactions might alter treatment differences. 
Finally, the feedback structure of the PGG introduced attribution ambiguity and signal dilution. Participants observed the aggregate group contribution after each round, but not the individual contributions of specific group members. As a result, changes in the group total could not be attributed to the AI-labelled player, to a specific human player, or to general variation within the group. Moreover, the bot represented only one of four group members, meaning that even an extreme strategy, such as full free-riding or full cooperation, was reflected in the feedback signal as only one quarter of the group's total contribution. The absence of strong bot-strategy or label effects should therefore not be interpreted as evidence that participants would respond similarly when AI behaviour is individually visible and attributable. Rather, our findings show that the AI label and scripted strategy did not produce detectable divergence under anonymous aggregate-feedback conditions. As discussed in Section 4.2, we regard this informational structure not merely as a limitation but as a defining boundary condition of the equivalence we report. Future work should compare aggregate and individual-feedback designs to test whether visible AI contributions elicit stronger differentiation, sanctioning, or strategic responses.

\bigskip

\section{Conclusion}
This study examined how cooperative norms function in hybrid human–AI groups. Using a repeated Public Goods Game followed by a one-shot Prisoner’s Dilemma, we found that cooperation patterns and normative expectations were statistically indistinguishable whether one group member was labelled as human or as AI, with contributions formally equivalent within ±5 tokens. These results indicate a form of bounded normative equivalence where the behavioural patterns that sustain cooperation, such as reciprocity, conditionality, and responsiveness to group behaviour, appeared unchanged when artificial agents were introduced under the conditions studied here. Rather than demonstrating algorithm aversion or moral disengagement, our findings highlight the stability of cooperative norms in hybrid groups. When feedback is anonymous and aggregated, individuals appear to rely on shared group signals rather than categorical distinctions between humans and AI. This baseline of normative equivalence provides a foundation for future research examining when such stability persists and when it breaks. This is particularly true in richer, more communicative, or more adaptive human–AI interactions, where moral agency, responsibility, trust, and transparency must be negotiated more actively.
\bigskip

\section*{CRediT authorship contribution statement}
Nico Mutzner: Conceptualization, Methodology, Formal analysis, Writing - Original Draft, Writing - Review \& Editing.\newline
Taha Yasseri: Conceptualization, Methodology, Resources, Writing - Review \& Editing, Supervision. \newline
Heiko Rauhut: Methodology, Resources, Writing - Review \& Editing, Supervision, Funding acquisition. \newline

\section*{Declaration of Generative AI and AI-assisted technologies in the writing process}
The authors used ChatGPT (4o/5), Claude Code (Opus 4.8) \& Gemini 3 for (a) language-related tasks such as proofreading, improving grammar and style, rephrasing sentences for clarity, and translation, and (b) for optimizing code used in statistical analyses (e.g., syntax correction, code efficiency, formatting). After using this tool, the authors reviewed and edited the content as needed and take full responsibility for the article's content.

\section*{Declaration of competing interest}
The authors declare that they have no known competing financial interests or personal relationships that could have appeared to influence the work reported in this paper.

\section*{Funding}
This work was supported by Grant BSSGI0\_155981 and Grant 10001A\_176333/1 from the Swiss National Science Foundation.\newline
Taha Yasseri was partially funded by Research Ireland under grant number IRCLA/2022/3217, ANNETTE (Artificial Intelligence Enhanced Collective Intelligence). TY also thanks Workday Inc. for financial support. 

\section*{Data availability}
The anonymized data, analysis scripts, and study materials needed to reproduce the reported analyses will be made available in a public repository upon acceptance for publication.

\newpage
\bibliographystyle{apalike}
\bibliography{references_linked}

\newpage
\appendix

\section{Appendix 1: Study Instruments}
  % Requires: \usepackage{booktabs, tabularx, threeparttable}
  \begin{table}[htbp]
  \centering
  \caption{Summary of study instruments and measures}
  \label{tab:instruments}
  \begin{threeparttable}
  \renewcommand{\arraystretch}{1.3}
  \small
  \begin{tabularx}{\textwidth}{@{}l X c l@{}}
  \toprule
  \textbf{Instrument / Stage} & \textbf{Construct(s) measured} & \textbf{Items} & \textbf{Response format} \\
  \midrule

  Public Goods Game (PGG) &
  Cooperative contribution to a shared public good; teammate framing (Human vs.\ AI); manipulation/attention check &
  10 rounds &
  0--100 tokens\tnote{a} \\
  \addlinespace

  Prisoner's Dilemma (PD) &
  One-shot cooperation/defection; expectation of partner's choice &
  2 &
  Cooperate / Defect \\
  \addlinespace

  Norm elicitation\tnote{b} &
  Social appropriateness of contribution levels (0, 25, 50, 75, 100 tokens) &
  5 &
  4-pt appropriateness \\
  \addlinespace

  Empirical \& normative expectations &
  Beliefs about others' average contribution and about what one \emph{should} contribute &
  2 &
  Slider, 0--100 \\
  \addlinespace

  Post-survey: general block &
  Trust\tnote{c}; perceived fairness; group cohesion; normative pressure &
  8 &
  6-pt Likert \\
  \addlinespace

  Post-survey: AI block\tnote{d} &
  Trust in AI; social inclusion of AI; tool-vs-teammate perception; attribution/blame; AI delegation; fairness of AI &
  8 &
  6-pt Likert \\
  \addlinespace

  \bottomrule
  \end{tabularx}
  \begin{tablenotes}[flushleft]
  \footnotesize
  \item[a] Linear PGG; endowment 100 tokens, multiplier 1.5, pooled and divided among 4 group members (3 human participants + 1 bot). Bot strategies: always cooperate, match the group average, or never cooperate.
  \item[b] Krupka--Weber-style social-appropriateness elicitation \citep{krupkaIdentifyingSocialNorms2013}.
  \item[c] Trust items adapted from \citep{kulmsMoreHumanLikenessMore2019}.
  \item[d] Administered only to participants in the AI treatment.
  \end{tablenotes}
  \end{threeparttable}
  \end{table}

\section{Appendix 2: Sociodemographics}

\begin{table}[H]
\centering
\caption{Sociodemographic characteristics of participants (N = 236)\label{tab:sociodemographics}}
\begin{tabular}{lll}
\toprule
\textbf{Characteristic} & \textbf{Category} & \textbf{n (\%)} \\
\midrule
Age (years) & 18--27 & 67 (29.0) \\
            & 28--37 & 96 (40.7) \\
            & 38--47 & 33 (14.0) \\
            & 48--57 & 22 (9.4) \\
            & 58--67 & 11 (4.7) \\
            & 68--70 & 5 (2.1) \\
\addlinespace
Sex         & Female & 117 (49.6) \\
            & Male   & 116 (49.2) \\
            & Prefer not to say & 1 (0.4) \\
\addlinespace
Ethnicity   & White  & 111 (47.0) \\
            & Black  & 100 (42.4) \\
            & Mixed  & 11 (4.7) \\
            & Asian  & 10 (4.3) \\
            & Other  & 1 (0.4) \\
\addlinespace
Top 5 Nationalities & South Africa   & 84 (35.6) \\
                    & United Kingdom & 46 (19.5) \\
                    & United States  & 35 (14.8) \\
                    & Poland         & 12 (5.1) \\
                    & Kenya          & 8 (3.4) \\
\addlinespace
Student status & Yes & 70 (29.7) \\
               & No  & 132 (55.9) \\
               & Missing & 32 (13.6) \\
\addlinespace
Employment status & Full-time  & 131 (55.5) \\
                  & Part-time  & 37 (15.7) \\
                  & Unemployed & 15 (6.4) \\
                  & Other      & 22 (9.3) \\
                  & Missing    & 29 (12.3) \\
\bottomrule
\end{tabular}
\end{table}

\end{document}